\documentclass[twoside,leqno,twocolumn]{article}  
\usepackage{ltexpprt} 
\usepackage{amsmath}

\usepackage{tikz}
\usetikzlibrary{arrows, shapes, trees, backgrounds}

\usepackage{floatrow}
\DeclareFloatFont{small}{\small}% "scriptsize" is defined by floatrow, "tiny" not
\begin{document}

\title{\Large Dropout Training of Matrix Factorization and Autoencoder for Link Prediction in Sparse Graphs}
\author{Shuangfei Zhai \thanks{Binghamton Univeristy, USA. szhai2@binghamton.edu}
\and Zhongfei (Mark) Zhang\thanks{Binghamton University, USA. zhongfei@cs.binghamton.edu}}
\date{}

\maketitle

%\pagenumbering{arabic}
%\setcounter{page}{1}%Leave this line commented out.

\begin{abstract} 
\small\baselineskip=9pt 
Matrix factorization (MF) and Autoencoder (AE) are among the most successful approaches of unsupervised learning. While MF based models have been extensively exploited in the graph modeling and link prediction literature, the AE family has not gained much attention. In this paper we investigate both MF and AE's application to the link prediction problem in sparse graphs. We show the connection between AE and MF from the perspective of multiview learning, and further propose MF+AE: a model training MF and AE jointly with shared parameters. We apply dropout to training both the MF and AE parts, and show that it can significantly prevent overfitting by acting as an adaptive regularization. We conduct experiments on six real world sparse graph datasets, and show that MF+AE consistently outperforms the competing methods, especially on datasets that demonstrate strong non-cohesive structures.
\end{abstract}

\section{Introduction}  

Link prediction is one of the fundamental problems of network analysis, as pointed out in \cite{kleinberg},  \textit{"a network model is useful to the extent that it can support meaningful inferences from observed network data." }   Given a graph $G(V, E)$ together with its adjacency matrix $A \in \{0, 1\}^{N \times N}$,  the set of nodes $V$, and the set of edges $E$, link prediction can be considered as a matrix completion problem on $A$. The problem is challenging because $A$ is often large and sparse,   which means that only a small fraction of the links are observed. As a result, a good model should have enough capacity to accommodate the complex connectivity pattern between all $N^2$ pairs of nodes, as well as strong generalization ability to make accurate predictions on unobserved pairs.

Among the large number of models proposed over the decade, Matrix Factorization (MF) is one of the most popular ones in network modeling  and relational learning in general \cite{relational, netflix, matfact, bigclam}. In its simplest form, MF directly models the interaction between a pair of nodes as the inner product of two latent factors, one for each node. This assumes that the large square adjacency matrix $A$  can be factorized into the product of a tall, thin matrix and a short, wide matrix as $A \approx W_2 W_1$, where $W_2 \in R^{N \times K}, W_1 \in R^{K \times N}$.  Each row of $W_1$ and each column of $W_1$ are often called a latent factor, as they often capture the community membership information of the nodes. Training of such a model is usually conducted with stochastic gradient descent, which makes it easily scalable to large datasets \cite{relational, netflix}.

Bayesian models, such as MMSB \cite{mmsb, jordan, bmf} are another family of latent factor models for studying network structures.  Compared with MF which directly learns the  latent factors as free parameters by solving an optimization  problem, Bayesian models treat the latent factors as random variables and model the stochastic process of the creation of a link. As a result, they can be considered as a stochastic version of MF. By putting various priors on the latent factors, link prediction is reduced to the inference problem of the posterior distribution of the link status. While Bayesian models can often significantly reduce overfitting compared with their deterministic counterparts, the inference processes such as MCMC and Variational Inference are usually much slower to run than direct optimization. As a result, their application has been limited to only moderate sized datasets. 

Autoencoder (AE) together with its variants such as Restricted Boltzman Machine (RBM) has recently achieved great success in various machine learning applications, and is well recognized as the building block of Deep Learning \cite{deeplearning, hinton2006}. AE learns useful representations by learning a mapping from an input to itself, which makes it different  from the above mentioned approaches. Surprisingly, it is not until recently that AE finds its application to  modeling graphs \cite{mdm, sdm14rbm}. In this paper we investigate the effectiveness of AE on the link prediction problem. In particular, we show that AE is closely connected to MF when unified in the same architecture. Motivated by this observation, we argue that MF and AE are indeed two complementary views of the same problem, and propose a novel model MF+AE where MF and AE are jointly trained with shared parameters.

To prevent overfitting, we train the model with Dropout \cite{dropout} combined with stochastic gradient descent. We highlight the effect of dropout training to MF+AE, and show that when approximated by the second order Taylor expansion,  the dropout training effectively penalizes a scaled $\ell_2$ norm of  (the combination of) the rows or columns of the weight matrices.  We evaluate MF+AE on six real world sparse graphs and show that dropout significantly mitigates overfitting on both MF and  AE, and that MF+AE outperforms the competing methods on all the datasets.

\section{Model}

\textbf{Matrix Factorization:}
Given an undirected graph $G(V, E)$ with $N$ nodes and $|E|$ edges, MF approximates its adjacency matrix $A \in \{0, 1\}^{N \times N}$ with the product of two low rank matrices $W_2$ and $W_1$ by solving the optimization problem:
\begin{equation} \label{eq:matfact}
\begin{split}
	&\min \sum_{i, j}{ L(A_{i, j}, \; g(W_2^j W_{1, i} + b_{1, i} + b_{2, j}))} 
\end{split}
\end{equation} where $W_1 \in R^{K \times N}, W_2 \in R^{N \times K}$; $b_1 \in R^N, b_2 \in R^N$ are the biases, $W_{1,i}$ is the $i^{th}$ column of $W_1$, $W_2^j$ is the $j^{th}$ row of $W_2$, $g$ is the link function,  $L$ is the loss function defined on each pair of nodes, and $K$ is the dimension of the latent factor. Since $A$ is symmetric for undirected graph, it is sometimes useful to adopt tied weights, i.e., $W_1 = W_2, b_1 = b_2$. We refer to the model with tied weights as the symmetric version of MF in this paper.

\textbf{Autoencoder:}
AE is originally proposed as an unsupervised representation learning method.  Given an example represented by a feature vector $x$,  an AE learns a reconstruction of itself by a function $\tilde{x} = F(x)$. The typical form of $F(x)$ can be characterized by a neural network with one hidden layer, where $x$ is first mapped to a hidden layer representation $h$, then a reconstruction $\tilde{x}$ is obtained by mapping $h$ to the original feature space:

\begin{equation}
\begin{split}
h  &= f(W_1 x + b_1) \\
\tilde{x} &= g(W_2 h + b_2)
\end{split}
\end{equation}with $W_1 \in R^{K\times N}, b_1 \in R^K, W_2 \in R^{N \times K}, b_2 \in R^N$. The parameters are learned by solving the following optimization problem:
\begin{equation} \label{eq:aeplain}
\begin{split}
\min \sum_i{L(x_i, \tilde{x}_i; W_1, b_1, W_2, b_2)}
\end{split}
\end{equation} 
Here we have slightly overloaded the loss function $L$ by defining it on the column vector $x_i$. The natural way of applying AE to modeling graphs is to represent each node as the set of its neighbors; in other words, set $x_i  = A_i$. This is analogous to the bag of  words representation prevalent in the document modeling community, and we call it the bag of nodes representation. Note that this representation is sufficient since when only the topological structure is available, we can learn an unseen node if we know all its neighbors. 

\textbf{The Joint Model:}
To better see the connection and difference between MF and AE, we now rewrite \eqref{eq:aeplain} by substituting $x_i$ with $A_i$:

\begin{equation} \label{eq:ae2}
\begin{split}
&\min \sum_{i, j} L(A_{i,j}, g(W_2^j h_i + b_{2,j}))\\
&s.t. \; h_i = f(W_1 A_i + b_1)
\end{split}
\end{equation}
And we rewrite \eqref{eq:matfact} by omitting the $b_1$term:

\begin{equation} \label{eq:matfact2}
\begin{split}
&\min \sum_{i,j} L(A_{i, j}, g(W_2^j h_i + b_{2, j})) \\
&s.t. \; h_i = W_1 \delta_i
\end{split}
\end{equation}where $\delta_i \in R^N$ is the indicator vector of node $i$, which is a binary vector with $1$ at the $i^{th}$ entry and $0$ for all the rest. We deliberately organize \eqref{eq:ae2} and \eqref{eq:matfact2} in such a way that in the high level, they share the same architecture. Both models first learn a hidden representation $h_i$, which is then fed through a classifier with link function $g$ and loss function $L$. The main difference is only in the form of the hidden representation.  For each node $i$, MF only looks at its id, and the hidden layer representation is learned by simply extracting the $i^{th}$ column of $W_1$. For AE, we first sum up the columns of $W_1$ indicated by $i$'s neighbors, and then pass the sum through an activation function $f$.  As a result, in MF, two nodes propagate "positive" information to each other only if they are directly connected; in AE, however, two nodes can do so as long as they appear in the same neighborhood of some other node, even if they are not directly connected. The different ways of the information propagation between that two models indicates that MF and AE are complementary  to each other to model different aspects of the same topological structure. 

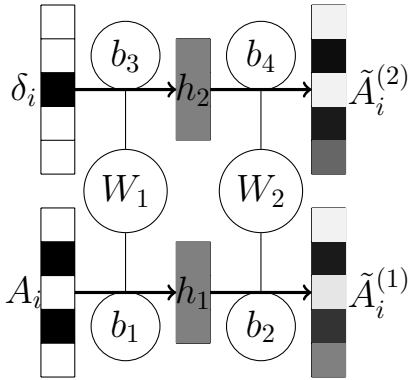
\begin{figure}[h] 
\centering
\begin{tikzpicture}[set style={{help lines}+=[dashed]}, scale = 0.45]
%\draw[style=help lines] (0,0) grid +(9,11);
\draw           (0,0) grid +(1,5);
\draw			(0,6) grid +(1,5);
\fill[black] (0, 1) rectangle +(1,1);
\fill[black] (0, 3) rectangle +(1, 1);
\fill[black] (0, 8) rectangle +(1, 1);

\draw  (4,  1) grid + (1,3);
\draw (4, 7) grid +(1, 3);
\fill[gray] (4, 1) rectangle +(1,3);
\fill[gray] (4, 7) rectangle +(1,3);

\draw    (8,0) grid +(1,5);
\draw (8, 6) grid +(1,5);

\fill[black!80] (8, 1) rectangle +(1,1);
\fill[black!90] (8, 3) rectangle +(1, 1);
\fill[black!50] (8, 0) rectangle + (1, 1);
\fill[black!10] (8, 2) rectangle + (1, 1);
\fill[black!5] (8, 4) rectangle + (1, 1);

\fill[black!90] (8, 7) rectangle +(1,1);
\fill[black!95] (8, 9) rectangle +(1, 1);
\fill[black!60] (8, 6) rectangle + (1, 1);
\fill[black!5] (8, 8) rectangle + (1, 1);
\fill[black!5] (8, 10) rectangle + (1, 1);

\draw [black, very thick, ->] (1, 8.5) -- (4, 8.5);
\draw [black, very thick, ->] (1, 2.5) -- (4, 2.5);
\draw [black, very thick, ->] (5, 8.5) -- (8, 8.5);
\draw [black, very thick, ->] (5, 2.5) -- (8, 2.5);

\node[circle, draw] (w1) at (2.5, 5.5) {\Large $W_1$};
\node[circle, draw]  (w2) at  (6.5,5.5) {\Large$W_2$};
\node[circle, draw] (b3) at (2.5, 9.5) {\Large $b_3$};
\node[circle, draw] (b1) at(2.5, 1.5) {\Large $b_1$};
\node[circle, draw] (b2) at(6.5, 1.5) {\Large $b_2$};
\node[circle, draw] (b4) at(6.5, 9.5) {\Large $b_4$};

\draw[black, -] (w1) -- (b3);
\draw[black, -] (w1) -- (b1);
\draw[black, -] (w2) -- (b2);
\draw[black, -] (w2) -- (b4);

\node at (-0.5, 8.5) {\Large $\delta_i$};
\node at (-0.5, 2.5) {\Large $A_i$};
\node at (4.5, 8.5) {\Large $h_{2}$};
\node at (4.5, 2.5) {\Large $h_{1}$};
\node at (10, 2.5) {\Large $\tilde{A}^{(1)}_{i}$};
\node at(10, 8.5) {\Large $\tilde{A}^{(2)}_{i}$};
\end{tikzpicture}
\caption{The architecture of the MF+AE model. The top part corresponds to the MF module, where the input is the indicator vector $\delta_i$; the bottom part corresponds to the AE module, where the input is the set of neighbors represented by the vector $A_i$; The two modules share the same transform matrices $W_1$ and $W_2$. The model is trained such that both of the two views can reconstruct the adjacency matrix $A$ .}
\label{fig:composite}
\end{figure}

We can also interpret the distinction between MF and AE as  two different views of the same entity: MF uses $\delta_i$, and AE uses $A_i$. We note that the two views are disjoint and sufficient: they do not overlap with each other, but each of them can sufficiently represent a node.  This perspective motivates  us to build a unified architecture where we train the two models jointly, and require both of them to be able to uncover the graph structure. The idea is similar to co-training \cite{cotraining} in the semisupervised learning setting, where one trains two classifiers on two sufficient views such that the two views can "teach" each other on unlabeled data. While in our problem, there is no "unlabeled data" available, we argue that the model can still benefit from the co-training idea  by requiring the two views to "agree with" each other. We then formulate our model, which we call MF+AE, as follows:

\begin{equation} \label{eq:composite}
\begin{split}
\min &\sum_{i }{L(A_{i}, g ( W_2 h_{1, i} + b_2))} \\
&+\rho \sum_{i }{L(A_{i}, g (W_2h_{2, i} + b_4))} \\
s.t.\; & h_{1, i} = f(W_1 A_i + b_1), \\
&h_{2, i} = f(W_1 \delta_i + b_3)
\end{split}
\end{equation}
In \eqref{eq:composite}, the objective function is composed of two parts, one for AE and the other for MF. $h_{1,i}$ and $h_{2, i}$ are the hidden representation learned by the AE and MF part, respectively, and the "agreement" is achieved by using the same set of weights $W_1$ and $W_2$ for both AE and MF. We modify the architecture of the MF objective by adding the same activation function $f$ and a corresponding bias term $b_3$. $\rho$ is a positive real number which could be simply set to $1$ in practice. We show the architecture of MF+AE in Figure \ref{fig:composite}.

In Figure \ref{fig:composite}, the bottom part and top part correspond to the first and second line of \eqref{eq:composite}, respectively. The color of each entry of the vector indicates its value, with white for $0$ and black for $1$. We see that the MF module is trained to reconstruct the neighborhood structure based on $\delta_i$, while the AE module learns to reconstruct $A_i$ from itself. The reconstructions from AE and MF are denoted as $\tilde{A}^{(1)}_{i}$ and $\tilde{A}^{(2)}_{i}$, respectively. Note that although the $5^{th}$ node does not appear in the neighbor of $i$ ($A_{i, 5} = 0$), the reconstruction is close to $1$ (dark gray). This means that we can make a confident prediction of the link of $i$ to node $5$.
 After the two reconstructions are obtained, the final prediction is calculated as the geometric mean of the two:
\begin{equation}
\tilde{A}_i = \sqrt[1 + \rho]{\tilde{A}^{(1)}_{i}  \odot (\tilde{A}^{(2)}_{i})^{\rho}}
\end{equation}
where $\odot$ denotes the element-wise product of two vectors.

\textbf{Activation Function and Loss Function:}
For the activation function $f$, we choose the Rectified Linear Units ($ReLU$), which is defined as:  
\begin{equation}
f(x) = \max(0, x)
\end{equation}
$ReLU$ provides nonlinearity by zeroing out all the negative inputs and keeping positive ones intact. We choose $ReLU$ over other popular choices such as $Sigmoid$ or $\tanh$ for two reasons. First, it does not squash the output to a fixed interval as $Sigmoid$ or $tanh$ does. As a result, it is closest to our intuition of approximating the behavior of MF. In fact, from the point of view of MF, the effect of $ReLU$ can be considered as putting a non-negativity constraint on $h$, which is closely related to the idea of Non-negative Matrix Factorization. Secondly, $ReLU$ is fast to compute compared with its alternatives, and still provides enough nonlinearity to significantly improve the model capacity over linear structures.

For $g$ and $L$, we choose the $Sigmoid$ function combined with cross entropy loss: 
\begin{align}
g(x) = \frac{1}{1 + e^{-x}}  
\end{align}
\begin{equation}
L(x, \tilde{x}) = -x\log{\tilde{x}} - (1-x)\log(1 - \tilde{x})
\end{equation}

The saturating property of the $Sigmoid$ function endows the model much flexibility since $h_{1, i}$ and $h_{2, i}$ need only to have similar activation patterns to both achieve good reconstructions; cross entropy is naturally a more appropriate choice than square loss for binary matrix.  Moreover, as A is often extremely sparse, reconstructing the whole matrix incurs the class imbalance problem, which means that the loss of the positive entries is dominated by the negative entries. As a result, it is important to reweight the cost of the positive and negative classes by utilizing the cost sensitive strategy. Consequently, our final form of the loss function becomes: 

\begin{equation}
L(A_i, \tilde{A}_i) = \sum_{j \approx i}{-\log \tilde{A}_{i, j} - \eta \sum_{j \not\approx i}\log (1 - \tilde{A}_{i, j})}
\end{equation}where $\tilde{A}_i = g(W_2 h_i + b_2)$ is the reconstruction of $A_i$; $j \approx {i}$ if $A_{i, j} = 1$, otherwise $j \not\approx i$. In practice, it is sufficient to approximate the second part of $L$ by a few samples; that is to say that at each training iteration we only need to sample part of the non-links for each node. Doing this may greatly speed up the training process on large graphs without sacrificing the performance. $\eta$  is the weight for the loss of the negative entries, which can be simply set as $\frac{\#nonlink \, samples}{\#links}$. 

\textbf{Dropout As Regularization:}
Regularization is critical for most machine learning models to generalize well to unseen data. Instead of putting explicit constraints on the parameters or hidden units, we use dropout training \cite{dropout, percy} as an implicit regularization. Dropout is a technique originally proposed for training feed forward neural networks to avoid overfitting. It works as follows: in the stochastic gradient training, we randomly drop out half of the hidden units (for both AE and MF)  and half of the feature units (for AE) for each node in the current iteration.  Mathematically, the effect of  dropping out can be simulated as applying an element-wise random dropout mask as follows:

\begin{equation} \label{eq:dropout0}
\begin{split}
\min \sum_{i} &\mathrm{E}_{\xi_{h,i}, \xi_{in, i}}\{L(A_{i}, g(W_2(\xi_{h,i} \odot h_{1,i}) + b_2)) \\
& + L(A_{i}, g(W_2(\xi_{h,i} \odot h_{2, i}) + b_4)) \}\\
s.t. \; & h_{1, i} = f(W_1(\xi_{in,i} \odot A_i) + b_1), \\
& h_{2, i} = f(W_1 \delta_i + b_3)
\end{split}
\end{equation}
Here $\xi_{h,i} \in \{0,1\}^K$ and  $\xi_{in,i} \in \{0, 1\}^N$ are the dropout masks for the hidden and input units, respectively; each element of them is an iid draw from the Bernoulli distribution. And $\odot$ is the element-wise product of two vectors. Note that we use the same dropout mask $\xi_{h,i}$ for both the AE and MF modules. This is to ensure that dropout does not cause any difference in the architecture between the two views.

For the AE module, randomly dropping out the input can be considered as a "denoising" technique, which was exploited by the previous work \cite{sda, chen2014}, and also was applied to link prediction \cite{mdm}. The motivation is that a model should be able to make a good reconstruction under a noisy or partial input. This property is particularly interesting to our problem because this is exactly the same link prediction problem: prediction of the whole based on parts. 

While theoretically explaining the effect of dropout is difficult for complex models, we can still gain an insight by looking at an approximate surrogate. Previously, \cite{percy, chen2014} used the second order Taylor expansion to explain the effect of feature noising in generalized linear models and AE, respectively. We can borrow the same tool to showcase a simplified version of MF+AE.

\textbf{Dropout for Matrix Factorization:} We consider the effect of dropout training on \eqref{eq:matfact}. For a concise articulation we ignore the bias terms;  the resulting model is described as the following objective function: 

\begin{equation} \label{eq:mfdrop}
O =  \sum_{i, j} \mathrm{E}_{\xi_{i}}\{L(A_{i, j}, g(W_2^j ( \xi_i \odot W_{1,i} ))\}
\end{equation} When $Sigmoid$ activation function with the cross entropy loss is used, we compute the second order approximation of \eqref{eq:mfdrop} in a closed form as:

\begin{equation} \label{eq:approxmf}
\begin{split}
\tilde{O} =\sum_{i,j} &L(A_{i, j}, g(\frac{1}{2}W_2^j W_{1,i} )) \\
&+\mathrm{E}_{\xi_{i}}\{(W_2^j \odot W_{1, i}^T)(\xi_{i} - \frac{1}{2}e)(g_{i,j} - A_{i,j})\}\\
&+\frac{1}{2}\mathrm{E}_{\xi_{i}}\{((W_2^j \odot W_{1, i}^T)(\xi_{i} - \frac{1}{2}e))^2g_{i,j}(1 - g_{i,j})\}\\
= \sum_{i,j} &L(A_{i, j}, g(\frac{1}{2}W_2^j W_{1,i} ))\\
&+ \frac{1}{8}(W_2^j)^2(W_{1,i})^2g_{i,j}(1 - g_{i,j})\\
 = \sum_{i} & L(A_{i}, g(\frac{1}{2}W_2 W_{1, i})) \\
 & +\frac{1}{8} \underbrace{(\sum_j (W_2^j)^2g_{i,j}(1-g_{i,j}))}_\text{$\lambda^{i}$}(W_{1,i})^2 
\end{split}
\end{equation}
where $e$ is a column vector of all $1$s, and $(W_2^j)^2$ and $(W_{1,i})^2$ are the element-wise square of the row and column vectors, respectively; $g_{i,j}$ is short for $g(\frac{1}{2}W_2^j W_{1,i} )$. The first equality of \eqref{eq:approxmf} is the result of the second order Taylor expansion; the second equality performs the expectation over the random variable $\xi_{i}$ whose $K$ entries are iid Bernoulli variables; the third equality is just a reorganization. We see that with the second order approximation, the dropout effect can be split into two factors. The first term is equivalent to the original objective except that the activation of each pair is scaled down by a half. The second part is more interesting; it can be considered as the product of a row vector $\lambda^{i}$ and the square of the column vector of $W_1$. Note that if we set $\lambda^{i} \propto e$ , the second term is reduced to the ordinary $\ell_2$ regularization on each column of $W_1$. In the case of dropout, however, $\lambda^{i}$ is equivalent to a weighted sum of the square of the rows of $W_2$, where the weight of each row of $W_2$ is determined by the degree of uncertainty of the prediction $g_{i,j}$. The overall effect of this regularization is two folds. First, it encourages the model to make confident predictions everywhere by minimizing $g_{i,j}(1 - g_{i, j})$; secondly, it performs a scaled version of $\ell_2$ regularization on the columns of $W_1$: coordinates that are highly active in the rows of $W_2$ are less penalized in the columns of $W_1$, and vice versa. In other words, the penalization on the column vectors of $W_1$ is adapted both to the structure of $W_2$ and the uncertainty of the predictions. This is in stark contrast to $\ell_2$ regularization where the penalization is uniformly put on each column of $W_1$. Finally, note that since the roles of $W_1$ and $W_2$ are exchangeable, the discussion of the regularization on the columns of $W_1$ also applies to the rows of $W_2$ by symmetry.

\textbf{Dropout for Autoencoder:} The nonlinear and nonsmooth nature of the ReLU activation function makes it difficult to analyze the behavior of dropout. We thus only show the case when $f$ is set as the identity function. Unsurprisingly, the effect of dropping out the hidden layer is similar to that of MF; the only difference is that we replace $W_{1, i}$ in \eqref{eq:approxmf} with $W_1 A_i$ . Following similar reasoning, it is obvious to see that dropping out the hidden layer in AE again penalizes the scaled $\ell_2$ norm of rows of $W_2$ in the same way. Its effect on  $W_1$ is more subtle:  instead of penalizing the norms of the columns of $W_1$ directly, the regularization is performed on the linear combinations of them. 

Next we proceed to study the effect of dropping out the input. Let us now denote $W = W_2W_1$ , and we have the dropout version of objective function:
\begin{equation}
O =  \sum_i \mathrm{E}_{\xi_i}\{ L(A_i, g(W(\xi_i \odot A_i) ))\}
\end{equation}
The second order approximation immediately follows as:
\begin{equation}
\begin{split}
\tilde{O} = \sum_{i } &L(A_i, g(\frac{1}{2}W A_i )) \\
&+\sum_j \mathrm{E}_{\xi_i}\{(W^j \odot A_i^T)(\xi_i - \frac{1}{2}e)( g_{i,j} - A_{i,j})\} \\
& + \frac{1}{2}\sum_j \mathrm{E}_{\xi_i} \{((W^j \odot A_i^T)(\xi_i - \frac{1}{2}e))^2g_{i,j}( 1-g_{i,j} )\} \\
= \sum_{i } &L(A_i, g(\frac{1}{2}W A_i )) \\
+\frac{1}{8} &\sum_j (W^j)^2\underbrace{\sum_i g_{i,j}(1-g_{i,j})(A_{i})^2 }_\text{$\lambda_{j}$}
\end{split}
\end{equation}
where $W^j = W_2^jW_1$ is the $j^{th}$ row of $W$, $g_{i,j}$ is short for $g(\frac{1}{2}W^jA_i)$. Recall that dropping out the hidden units in both MF and AE performs a scaled $\ell_2$ norm regularization on (the linear combinations of) \textbf{columns} of $W_1$; dropping out the input performs a scaled $\ell_2$ regularization on the linear combinations of \textbf{rows} of $W_1$. The regularization on $W_2$ is also very different from the case of dropping out the hidden units, but in a less clear way.

To summarize, we show that in the simplified case, dropping out the hidden units and inputs can be both interpreted as an adaptive regularization. They both push the model to make confident predictions by minimizing the factor $g_{i,j}(1 - g_{i,j})$, while they penalize different aspects of the parameter matrices. When combined in the joint training architecture, they provide complementary regularization to prevent MF+AE from overfitting.

\section{Experiments}
\subsection{Experiment Setup}
We conduct the experiments on six real world datasets: DBLP, Facebook, Youtube, Twitter, GooglePlus, and LiveJournal, all of which are available to download at \textit{snap.stanford.edu/data}. We summarize the statistics of the six datasets in Table \ref{tb:dataset}. Except for DBLP which is an author collaboration network, all the rest are online social networks. In particular, Youtube, Twitter, Gplus, and LiveJournal are all originally directed network; we convert them to undirected graphs by ignoring the direction of the links. 

\begin{table}[h] 
\centering
\begin{tabular}{|l|l|l|l|}
\hline
dataset     & N    & E       & D    \\ \hline
DBLP        & 2,958 & 64,674  & 21.9 \\ \hline
Facebook    & 2,277 & 148,546 & 65.2 \\ \hline
Youtube     & 1,955 & 102,950 & 52.6 \\ \hline
Twitter     & 2,477 & 107,895 & 43.6 \\ \hline
Gplus       & 2,129 & 148,306 & 69.7 \\ \hline
LiveJournal & 3,006 & 123,236  & 41.0 \\ \hline
\end{tabular}
\caption{Statistics of the datasets where N: number of nodes, E: number of links, D: average degree. }
\label{tb:dataset}
\end{table}
Following the experimental protocol suggested in \cite{randomwalk}, we first split the observed links into a training graph $G_{train}$ and a testing graph $G_{test}$. We then train the models on $G_{train}$, and evaluate the performance of the models on $G_{test}$. In particular, note that a naive algorithm which simply predicts a link for all the pairs that have at least one common neighbor would make a pretty good accuracy. We then only consider the nodes that are 2-hops from the target node as the candidate nodes \cite{randomwalk}. The metrics used are Precision at top 10 position($Prec@10$) and AUC. 

The methods we evaluate are:

\textbf{ Adamic-Adar Score (AA)} \cite{adamic}. This is a popular score based method, it calculates the pair-wise score of node $i, j$ as $S(i, j) = \sum \limits_{n \in CN(i, j)}\log(\frac{1}{d_n})$, where $CN(i,j)$ denotes the set of common neighbors of node $i$ and $j$, $d_n$ is the degree of node $n$. Prediction is made by ranking the score of all nodes to the target node.

\textbf{ Random Walk with restart (RW)}\cite{kleinberg}. RW uses the stationary distribution of a random walk from a given node to all the nodes as the score. The restart probability needs to be set in advance. In practice we find that while the performance of RW is pretty robust within a wide range of values, a relatively large value (compared with $0.3$ used in \cite{randomwalk}) works slightly better on our problem . We set it as $0.5$ throughout the experiments.

\textbf{Matrix Factorization with $\ell_2$ regularization (MF2)}. This is a variant of the model proposed in \cite{matfact}. To be fair for the comparison with the other models, we use the cross entropy loss instead of the rank loss proposed in the paper. We also use a weight decay of $10^{-5}$.

\textbf{Autoencoder with $\ell_2$ regularization (AE2)}. This is the model corresponding to \eqref{eq:aeplain} with an additional $\ell_2$ regularization on $W_1$ and  $W_2$. The weight decay parameter is also set as $10^{-5}$.

\textbf{Matrix Factorization with dropout (MFd)}. This corresponds to the model described in \eqref{eq:mfdrop} with the additional bias vectors. No weight decay is used.

\textbf{Autoencoder with Dropout (AEd)}. This is the single AE with $ReLU$ activation function and the cross entropy loss trained by dropout. 

\textbf{Marginalized Denoising Model (MDM)}\cite{mdm}. This is one of the few existing AE based models where  a linear activation together with square loss is used. MDM marginalizes the dropout noise of the features during training, but no hidden layer dropout is adopted.  The model requires the input noise level to be set;  we set it as $0.9$ throughout the experiments.

\textbf{The Joint Model (MF+AE)}. This corresponds to the model described in \eqref{eq:dropout0}, where we jointly train MF and AE with shared weights using dropout.

All the above methods are implemented in $Matlab$. We use the authors' implementation for MDM, and use our own implementations for the rest models. Note that MF2, AE2, MDM, MFd, AEd, and MF+AE  all require the dimensionality of the latent factor or hidden layer $K$ as the input. To be fair for a comparison, we do not attempt to optimize this parameter for each model. Instead, we set it the same for all of the models on all the six datasets. In the experiments, we use $100$.

Another important aspect for both MF and AE models is the choice of a symmetry model vs. an asymmetry model, i.e., whether or not to set $W_1 = W_2^T$. We find that AE models are less sensitive to the characteristics of a dataset, and almost always benefit from the tied weights. The optimal choice for MF models is, however, extremely problem-dependent. In our experiments, we use tied weights for all AE based models including MF+AE on all the six datasets. For MF based models, we use tied weights on Facebook, Twitter, DBLP, LiverJournal, and untied weights on Youtube and Gplus.

In this paper, we are interested in evaluating the performance of different models on \textbf{sparse} graphs. In other words, we investigate how well a model generalizes given only a sparse training graph $G_{train}$.  To this end,  each of the six datasets is chosen as a relatively densely connected subgraph as in Table \ref{tb:dataset}. We then randomly take $10\%$ of links for training and use the rest for testing.  In this way, we train the model on a sparse graph, while still have enough held out links for testing. We train all the models (except AA and RW which require no training) to convergence, which means that we do \textbf{not} use a separate validation set to perform early stopping.  We do this for two reasons. First, in sparse graphs, splitting out a separate set for validation is expensive. Secondly and more importantly, we are interested in testing the generalization ability of each model. In practice we find that almost all the models we have tested benefit from a properly chosen early stopping point. However, this makes the results very difficult to interpret as it is difficult to evaluate the contribution of early stopping in different models.

\begin{table*}[t]
\centering
\begin{tabular}{|l|l|l|l|l|l|l|l|}
\hline
Model                   & Facebook         & Twitter          & Youtube       & Gplus          & DBLP            & LiveJournal      & Average\\ \hline
MF+AE & \textbf{0.58057} & \textbf{0.46693} & \textbf{0.33132} & \textbf{0.41277} & \textbf{0.32462} & \textbf{0.29027}  & \textbf{0.4011} \\ \hline
AEd   & 0.54643          & 0.44229          & 0.31769          & 0.40085          & 0.29942          & 0.28659 	&0.3822         \\ \hline
AE2   & 0.37748          & 0.2773           & 0.087839         & 0.17973          & 0.28308          & 0.1722      &0.2296     \\ \hline
MFd   & 0.46716          & 0.4041           & 0.23636          & 0.28956          & 0.29599          & 0.23958     &0.3221     \\ \hline
MF2   & 0.45216          & 0.39823          & 0.13842          & 0.24594          & 0.30735          & 0.21651     &0.2931     \\ \hline
MDM   & 0.54255          & 0.41304          & 0.23548          & 0.3149           & 0.30286          & 0.25415     &0.3438     \\ \hline
RW    & 0.53143          & 0.40647          & 0.15805          & 0.21685          & 0.27757          & 0.20524      &0.2993    \\ \hline
AA    & 0.47439          & 0.34576          & 0.13647          & 0.17523          & 0.23712          & 0.18247       &0.2586   \\ \hline
\end{tabular}
\end{table*}

\begin{table*}
\centering
\begin{tabular}{|l|l|l|l|l|l|l|l|}
\hline
Model                   & Facebook         & Twitter          & Youtube       & Gplus          & DBLP        & LiveJournal   &Average   \\ \hline
MF+AE & \textbf{0.9136} & \textbf{0.8192} & \textbf{0.75832} & \textbf{0.82262} & \textbf{0.80695} & 0.75779    &  \textbf{0.8131}    \\ \hline
AEd   & 0.89655         & 0.81006         & 0.74512          & 0.80808          & 0.79257          & \textbf{0.76216} &0.8024\\ \hline
AE2   & 0.69946         & 0.65219         & 0.53972          & 0.58406          & 0.71501          & 0.61556        &0.6343  \\ \hline
MFd   & 0.82891         & 0.74193         & 0.68857          & 0.754            & 0.77362          & 0.71683         &0.7506 \\ \hline
MF2   & 0.75104         & 0.69247         & 0.64594          & 0.68186          & 0.76708          & 0.68844       &0.7045   \\ \hline
MDM   & 0.89366         & 0.7882          & 0.66943          & 0.74716          & 0.79974          & 0.72491        &0.7705  \\ \hline
RW    & 0.88156         & 0.79524         & 0.74236          & 0.76542          & 0.78078          & 0.71717        &0.7804  \\ \hline
AA    & 0.70223         & 0.57279         & 0.47685          & 0.57406          & 0.74724          & 0.5681          &0.6069 \\ \hline

\end{tabular}
\caption{Performance of each model on each dataset. The top half of the table is $Prec@10$, bottom half is $AUC$.  The best performance of each metric\&dataset combination is highlighted with  \textbf{bold face}.}
 \label{tb:res}
\end{table*}

\textbf{MF+AE Achieves Best Performance}
We first list the results of the experiments in Table \ref{tb:res} for a comparison. Overall, we see that MF+AE has the best average performance. In particular, it achieves the best performance on all the six datasets evaluated by $Prec@10$, and on all but the LiveJournal dataset evaluated by AUC. This shows that the joint training of MF and AE consistently boosts the generalization ability without increasing the model size. AEd achieves the second best average performance evaluated by both metrics. MDM, as a variant of AE, achieves the third best performance on $Prec@10$ and fourth on AUC. This is reasonable since on the one hand, the utilization of feature noising improves the generalization ability, and on the other hand, the linear nature limits its ability to model complex graph structures, and also  due to the use of the square loss and ignorance of the class imbalance, the performance further deteriorates in sparse graphs.  One seemingly surprising result is that RW performs pretty well  despite of its simplicity.

\begin{figure*}[t]
\centering
\includegraphics[width=\textwidth]{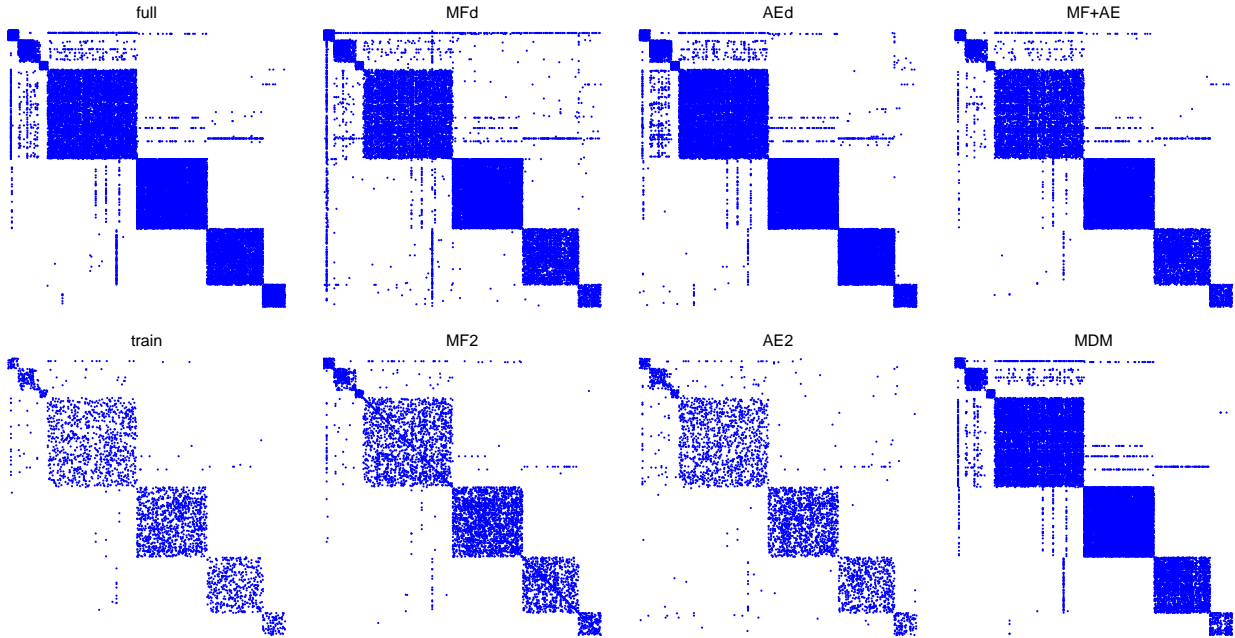}
\caption{Visualization of the Facebook dataset. From top left to bottom right: the full graph, the predictions of MFd, AEd, MF+AE, respectively, the training graph, the prediction of MF2, AE2, MDM, respectively.}
 \label{fg:facebook}
\end{figure*}

\textbf{Dropout Improves Generalization}
We note that for both AE and MF, the dropout version significantly outperforms their $\ell_2$ counterparts on both $Prec@10$ and AUC. Evaluated by the average performance,  AEd outperforms AE2 by $66\%$ on $Prec@10$, $26\%$ on AUC; MFd also outperforms MF2 by $10\%$ on $Prec@10$ and $6.5\%$ on AUC.  This verifies that dropout as an adaptive regularization performs much better than $\ell_2$ norm regularization, especially with AE whose objective function is highly nonconvex. To better understand this, we visualize the full graph, the training graph of the Facebook dataset together with the predictions made by each model in Figure \ref{fg:facebook}. We do not visualize the results of RW and AA since they do not output direct reconstructions. For each of the other six models, we convert the predictions to binary by a threshold at $0.5$. Also for a better visualization, we down sample all the graphs by $80\%$. 

In Figure \ref{fg:facebook} we see that AE2 and MF2  fit the training graph very well, but fail to uncover the much densely connected full graph.  However, with dropout training,  the predictions of AEd and MFd look much closer to the full graph than AE2 and MF2. This suggests that models trained with dropout generalize much better than their $\ell_2$ regularized counterparts.

We then compare the predictions of MF+AE, AEd, MFd, MDM, respectively, which all use (different variants of) dropout training. It is not difficult to see that MF+AE's prediction resembles the full graph the most. AEd and MFd make a lot of "False Positive" predictions which are clearly shown by the pepper salt like pixels off the diagonal. MDM makes more "False Negative" predictions, such as the missing of the small cluster near the right top corner. We note that the quality of the predictions shown by the visualization is also consistent with the results in Table \ref{tb:res}.

\textbf{Modeling Non-cohesive Graphs}
\begin{figure}[h!]
\centering
\includegraphics[scale = 0.45]{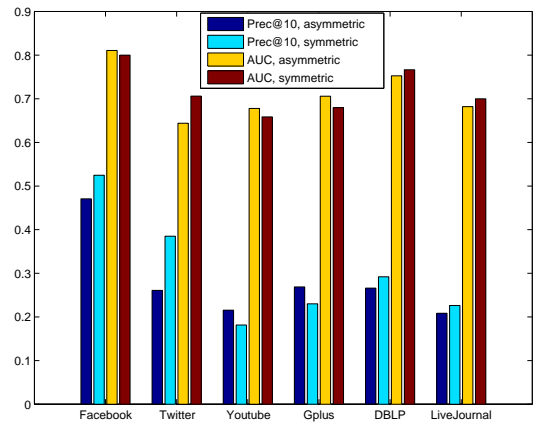}
\caption{The average performance of symmetric MF vs. asymmetric MF}
\label{fig:sym}
\end{figure}

Among all the six datasets we have tested, Youtube and Gplus datasets demonstrate least cohesive structures, i.e., they consist of more follower-followee relationships than the other datasets. The cohesive vs. non-cohesive distinction of graph structure has previously been investigated in \cite{2mode, hoff}. To show this,  we have trained the symmetric and asymmetric versions of MFd and MF2, respectively, on all the six datasets. We then report the averaged performances of the symmetric MF and asymmetric MF in Figure \ref{fig:sym}. We see that the symmetric version works better on Facebook, Twitter, DBLP, LiveJournal, and the asymmetric version works better on Youtube and Gplus. This experiment shows that Youtube and Gplus demonstrate more non-cohesive structure which cannot be symmetrically modeled by the inner product of two vectors. 

With this in mind, let us look at the performances of different models on these two datasets. It is clear that MF+AE and AEd as the best and second best models outperform the other methods by much larger margins than on the other four datasets. Note that AEd and MF+AE still use the tied weights on these two datasets, as we found little difference in performance when switched to untied weights. Also note that even though MDM uses feature dropout, it still fails to model the non-cohesive structures properly. We argue that it is the nonlinear activation function that gives the MF+AE and AEd more modeling power than linear models.

\section{Related Work}
The link prediction problem can be considered as a special case of relational learning and recommender systems \cite{relational, netflix}, and a lot of techniques proposed are directly applicable to link prediction as well. Salakhutdinov et al. \cite{rbmcf} first apply RBM to movie recommendation. Recently, Chen and Zhang \cite{mdm} propose a variant of linear AE with marginalized feature noise for link prediction, and Li et al. \cite{sdm14rbm} apply RBM to link prediction in dynamic graphs.

MF+AE is also related to the supervised learning based methods \cite{sdm06, perspectives}. While these approaches directly train a classifier on manually collected features, MF+AE directly learns the appropriate features from the adjacency matrix.

The utilization of dropout training as an implicit regularization also contrasts with Bayesian models \cite{mmsb, jordan}. While both dropout and Bayesian Inference are designed to reduce overfitting, their approaches are essentially orthogonal to each other. It would be an interesting future work to investigate whether they can be combined to further increase the generalization ability. Dropout has also been applied to training generalized linear models \cite{percy}, log linear models with structured output \cite{loglinear}, and distance metric learning \cite{dml}.

This work is also related to graph representation learning. Recently, Perozzi et al. \cite{deepwalk} propose to learn node embeddings by predicting the path of a random walk, and they show that the learned representation can boost the performance of the classification task on graph data. It would also be interesting to evaluate the effectiveness of MF+AE in the same setting.

\section{Conclusion}
We propose a novel model MF+AE which jointly trains MF and AE with shared parameters. We show that dropout can significantly improve the generalization ability of both MF and AE by acting as an adaptive regularization on the weight matrices. We conduct experiments on six real world sparse graphs, and show that MF+AE outperforms all the competing methods, especially on datasets with strong non-cohesive structures.

\section*{Acknowledgements}
This work is supported in part by NSF CCF-1017828, the National Basic Research Program of China (2012CB316400), and Zhejiang Provincial Engineering Center on Media Data Cloud Processing and
Analysis in China.

\end{document}